\documentclass[10pt,journal,compsoc]{IEEEtran}

\usepackage{cite}
\usepackage{amsmath,amssymb,amsfonts}
\usepackage{algorithmic}
\usepackage{algorithm}
\usepackage{graphicx}
\usepackage{textcomp}
\usepackage[table]{xcolor}
\usepackage{booktabs}
\usepackage{multirow}
\usepackage{enumitem}
\usepackage{url}
\usepackage{caption}
\usepackage{subcaption}
\usepackage{bm} 
\usepackage{tikz}
\usetikzlibrary{shapes.geometric, arrows, positioning, calc, backgrounds}

\definecolor{desat-orange}{RGB}{245, 205, 175}
\definecolor{desat-green}{RGB}{195, 230, 200}
\definecolor{desat-purple}{RGB}{215, 200, 235}
\definecolor{desat-blue}{RGB}{180, 215, 240}
\definecolor{desat-red-border}{RGB}{200, 80, 80}
\definecolor{desat-red-fill}{RGB}{245, 235, 235}
\definecolor{desat-gray}{RGB}{230, 230, 230}
\definecolor{border-gray}{RGB}{80, 80, 80}

\tikzstyle{startstop} = [rectangle, rounded corners=1mm, minimum width=2.5cm, minimum height=0.7cm, text centered, draw=border-gray, fill=white, font=\small, align=center]
\tikzstyle{io} = [trapezium, trapezium left angle=70, trapezium right angle=110, minimum width=2.0cm, minimum height=0.7cm, text centered, draw=border-gray, fill=desat-gray!70, font=\small\bfseries, align=center]
\tikzstyle{process} = [rectangle, minimum width=2.8cm, minimum height=0.9cm, text centered, draw=border-gray, fill=desat-blue, font=\small, text width=2.6cm, align=center]
\tikzstyle{agf_process} = [rectangle, minimum width=2.8cm, minimum height=0.9cm, text centered, dashed, draw=desat-red-border, fill=desat-red-fill, font=\small\bfseries, text width=2.6cm, rounded corners=1mm, align=center]
\tikzstyle{pruned_expert} = [circle, minimum width=1.6cm, text centered, draw=border-gray, fill=desat-green, font=\small, text width=1.4cm, align=center]
\tikzstyle{full_expert} = [circle, minimum width=1.6cm, text centered, draw=border-gray, fill=desat-orange, font=\small, text width=1.4cm, align=center]

\tikzstyle{decision} = [diamond, minimum width=2.0cm, minimum height=1.0cm, text centered, draw=border-gray, fill=desat-purple, font=\small\bfseries, inner sep=-1pt, align=center]

\tikzstyle{pareto} = [trapezium, trapezium left angle=110, trapezium right angle=70, minimum width=2.2cm, minimum height=0.8cm, text centered, draw=border-gray, fill=cyan!10, font=\small\bfseries, align=center]
\tikzstyle{arrow} = [thick,->,>=stealth, draw=border-gray]
\tikzstyle{dashed_arrow} = [thick,->,>=stealth, dashed, draw=border-gray]
\title{Alternating Gradient Flow Utility: A Unified Metric for Structural Pruning and Dynamic Routing in Deep Networks}

\author{Tianhao~Qian,
        Zhuoxuan~Li,
        Jinde~Cao, 
        Xinli~Shi,
        and~Leszek~Rutkowski 
\IEEEcompsocitemizethanks{
    \IEEEcompsocthanksitem T. Qian and J. Cao are with the School of Mathematics, Southeast University, Nanjing 210096, China (e-mail: qth2mir@seu.edu.cn; jdcao@seu.edu.cn; liuhanjie1993@gmail.com).
    \IEEEcompsocthanksitem Z. Li is with the School of Mathematics, Southeast University, Nanjing 210096, China, and also with the Systems Research Institute of the Polish Academy of Sciences, Warsaw 01-447, Poland (e-mail: 230229338@seu.edu.cn).
    \IEEEcompsocthanksitem X. Shi is with the School of Cyber Science \& Engineering, Southeast University, Nanjing 210096, China (e-mail: xinli\_shi@seu.edu.cn).
    \IEEEcompsocthanksitem L. Rutkowski is with the Systems Research Institute of the Polish Academy of Sciences, Warsaw 01-447, Poland; the Institute of Computer Science, AGH University of Krakow, Kraków 30-059, Poland; and also with SAN University, Łódź 90-113, Poland (e-mail: leszek.rutkowski@ibspan.waw.pl).
    \IEEEcompsocthanksitem J. Cao is the corresponding author.
}%
\thanks{}}

\begin{document}

\IEEEtitleabstractindextext{
\begin{abstract}
Efficient deep learning traditionally relies on static heuristics like weight magnitude or activation awareness (e.g., Wanda, RIA). While successful in unstructured settings, we observe a critical limitation when applying these metrics to the structural pruning of deep vision networks. These contemporary metrics suffer from a magnitude bias, failing to preserve critical functional pathways. To overcome this, we propose a decoupled kinetic paradigm inspired by Alternating Gradient Flow (AGF), utilizing an absolute feature-space Taylor expansion to accurately capture the network's structural `kinetic utility'. \textbf{First}, we uncover a topological phase transition at extreme sparsity, where AGF successfully preserves baseline functionality and exhibits topological implicit regularization, avoiding the collapse seen in models trained from scratch. \textbf{Second}, transitioning to architectures without strict structural priors, we reveal a phenomenon of Sparsity Bottleneck in \textbf{Vision Transformers (ViTs)}. Through a gradient-magnitude decoupling analysis, we discover that dynamic signals suffer from \emph{signal compression} in converged models, rendering them suboptimal for real-time routing. \textbf{Finally}, driven by these empirical constraints, we design a hybrid routing framework that decouples AGF-guided offline structural search from online execution via zero-cost physical priors. We validate our paradigm on large-scale benchmarks: under a 75\% compression stress test on \textbf{ImageNet-1K}, AGF effectively avoids the structural collapse where traditional metrics aggressively fall \emph{below} random sampling. Furthermore, when systematically deployed for dynamic inference on \textbf{ImageNet-100}, our hybrid approach achieves Pareto-optimal efficiency. It reduces the usage of the heavy expert by approximately \textbf{50\%} (achieving an estimated overall cost of 0.92$\times$) without sacrificing the full-model accuracy.
\end{abstract}

}

\maketitle

\section{Introduction}
\label{sec:introduction}

\IEEEPARstart{D}{eep} Neural Networks (DNNs) have evolved from static computational graphs to dynamic and flexible systems. To mitigate computational costs of modern vision backbones, two main paradigms have emerged: \textbf{Channel Pruning}, which permanently removes redundant structures \cite{han2015learning, liu2017learning}, and \textbf{Dynamic Routing}, which conditionally skips computations based on input complexity \cite{wang2018skipnet, huang2017multi}.

Beyond their success, a central challenge is the accurate estimation of channel importance. Standard approaches rely on magnitude-based heuristics (e.g., $\ell_1$-norm) \cite{li2016pruning} given that "smaller weights are less important." While effective in saturated settings, these static metrics often fail to capture the complex learning dynamics of modern networks. Conversely, gradient-based methods (e.g., SNIP, GraSP) \cite{lee2018snip, wang2020picking} utilize instantaneous sensitivity to estimate importance, often resulting in instability during optimization.

Recently, activation-aware pruning metrics (such as Wanda\cite{sun2024wanda} and RIA \cite{zhang2024plug}) have established themselves as highly effective criteria, mostly benefiting from their success in exploiting activation outliers in Large Language Models (LLMs). However, a critical vulnerability is revealed when these methods are applied to the structural pruning of deep vision backbones (e.g., ResNets). Our empirical results demonstrate a critical limitation of static metrics: at 25\% width retention, networks pruned by these metrics underperform even uniform random pruning, hitting an entropic lower bound. We identify this failure as a magnitude bias. Despite exhibiting low magnitudes, some neurons work as critical feature integrators in highly compressed routing pathways, where static metrics fail to preserve these vital topological pathways. While our primary rigorous analysis focuses on CNNs, we further demonstrate in our extended experiment(see Section \ref{sec:vit_experiments}) that this bias is equally destructive in architectures lacking strict structural priors, such as Vision Transformers (ViTs). Recent advancements, such as Wanda++\cite{yang2025wandaplus}, recognize this limitation by incorporating block-level regional gradients. Yet, these methods remain inherently confined to localized output matching and lack global topological awareness. In contrast, the end-to-end kinetic utility along the global optimization trajectory is captured via our framework.

To circumvent magnitude bias, our work is inspired by the Alternating Gradient Flows (AGF) framework \cite{kunin2025alternating}. This "gradient-driven utility" is reconstructed as a dynamic spatial criterion for structural efficiency. We define the utility $U(c)$ as the cumulative gradient norm, measuring the \emph{true learning potential} rather than static amplitude. In this way, AGF naturally acts as a topological proxy: it preserves low-magnitude neurons if their gradient flow indicates high sensitivity and contribution to the global loss reduction.

Translating this insight into practice reveals an asymmetry between topology construction and inference routing:
\begin{enumerate}
    \item \textbf{The Construction Phase (Pruning):} AGF Utility is highly effective at constructing robust network topologies. Capturing critical routing pathways allows the network to smoothly traverse the topological phase transition. Consequently, the model survives extreme structural pruning (e.g., 25\% width in ViTs, or 3\% width in ResNets)—conditions where magnitude-based selection simply fails to converge.
    
    \item \textbf{The Execution Phase (Routing):} AGF Utility is suboptimal as a real-time inference proxy. As the model converges, gradient dynamics diminish and saturate, offering little structural guidance. Instead, robust physical priors (like Confidence or $\ell_1$)  prevail for efficient execution.
\end{enumerate}

By resolving this dilemma, we propose a decoupled kinetic paradigm that uses AGF to \emph{build} the road (topology) and Confidence/$\ell_1$ to \emph{navigate} it (routing). In summary, our contributions follow a progressive trajectory from topological discovery to large-scale system deployment:

\begin{enumerate}
    \item \textbf{AGF-Guided Pruning and Topological Phase Transitions:} Through extensive analysis on CNN backbones, we identify a \emph{topological phase transition} at extreme sparsity boundaries. Under these constraints, networks trained from scratch inevitably collapse. In contrast, AGF safely bypasses magnitude bias by anchoring to orthogonal dynamic routing hubs, preserving structural functionality. Furthermore, we uncover a \emph{Topological Implicit Regularization} effect, demonstrating that stochastic gradient noise from sparse calibration significantly enhances the network's long-term recovery elasticity.
    
    \item \textbf{Signal Saturation and the ViT Sparsity Bottleneck:} Advancing beyond standard CNNs, we rigorously analyze metric fidelity across optimization trajectories. We reveal a severe \emph{signal compression} phenomenon in converged models, where the dynamic AGF signal dramatically underestimates the true physical cost ratio (compressing from $\sim 149.4\times$ to $\sim 21.4\times$). Consequently, when extending structural pruning to architectures lacking strict inductive biases like \textbf{Vision Transformers (ViTs)}, all deterministic metrics (both magnitude and gradient-based) hit a strict Sparsity Bottleneck. These dual limitations mathematically necessitate the decoupling of dynamic gradient proxies from static physical costs.
    
    \item \textbf{The Decoupled Kinetic Paradigm and Large-Scale Verification:} To resolve the aforementioned dilemma, we propose a hybrid routing framework that decouples offline topology construction (via AGF) from online dynamic execution (via zero-cost priors like Confidence). We subject this paradigm to an extreme 75\% structural compression stress test on \textbf{ImageNet-1K}, where traditional metrics aggressively destroy core routing pathways and fall \emph{below} the uniform random baseline. AGF, however, successfully navigates this intrinsic capacity limit. Furthermore, when deployed as a dynamic inference system on \textbf{ImageNet-100}, our input-adaptive approach successfully breaks the sparsity bottleneck. It matches the full-capacity baseline accuracy while reducing the usage of the heavy expert by $\approx \mathbf{50\%}$ (achieving an overall estimated computational cost of $\mathbf{0.92\times}$).
\end{enumerate}

\section{Related Work}
\label{sec:related_work}

This section locates our work at the intersection of feature learning dynamics, structural pruning (from CNNs to ViTs and LLMs), and dynamic routing.

\subsection{Theoretical Dynamics and Phase Transitions}
Recent years have witnessed empirical success of deep learning, but its optimization landscapes and information bottlenecks remain challenging. For traditional statistical learning bounds, it's quite struggling to explain the generalization of over-parameterized networks. 

Further, physics-inspired approaches have provided alternative perspectives. For example, studies on the Information Bottleneck theory \cite{shwartz2017opening} and optimization dynamics\cite{saxe2013exact, nakkiran2019deep} have revealed that neural networks undergo distinct topological phase transitions during training (e.g., grokking \cite{power2022grokking, liu2023omnigrok}). Concurrently, \textbf{Kunin et al. (2025)} \cite{kunin2025alternating} proposed the \textbf{Alternating Gradient Flow (AGF)} framework, modeling feature learning as a saddle-to-saddle transition driven by a specific kinetic utility function. Our work translates this macroscopic continuous theory into practical network design: the continuous AGF utility is repurposed into a discrete, $O(1)$-cost metric to overcome the severe entropic lower bounds typical of extreme structural compression.

\subsection{Neural Network Pruning and Magnitude Bias}
Pruning techniques trace back to \textbf{Optimal Brain Damage} \cite{lecun1989optimal} and \textbf{Optimal Brain Surgeon} \cite{hassibi1993second}, which harnessed second-order derivative information (e.g., WoodFisher \cite{singh2020woodfisher}). Modern approaches generally fall into two distinct trajectories:

\textbf{1) Magnitude and Activation-Aware Pruning:} Classical heuristics like $\ell_1$\cite{li2016pruning, he2017channel} assume small weights are redundant. Recently, activation-aware metrics have dominated the compression of Large Language Models (LLMs), with prominent methods like \textbf{Wanda} \cite{sun2024wanda}, \textbf{SparseGPT} \cite{frantar2023sparsegpt}, \textbf{LLM-Pruner} \cite{ma2023llm}, and \textbf{SliceGPT} \cite{ashkboos2024slicegpt} achieving state-of-the-art results by exploiting activation outliers \cite{yin2023outlier}. However, we identify a limitation: when applied to extreme structural pruning of vision networks (CNNs and ViTs), these static proxies suffer from a severe magnitude bias, often disproportionately eliminating critical low-magnitude structural pathways and degrading performance below random pruning baseline.

\textbf{2) Gradient-based Sensitivity:} Methods like \textbf{SNIP} \cite{lee2018snip}, \textbf{GraSP} \cite{wang2020picking}, and \textbf{SynFlow} \cite{tanaka2020pruning} leverage gradient signals to estimate structural importance \cite{wang2023unified}. \textbf{Taylor Pruning} \cite{molchanov2019importance} remains a cornerstone in this category. Yet, traditional first-order Taylor expansions suffer from signal cancellation and high variance. Furthermore, recent universal structural pruning frameworks (e.g., \textbf{DepGraph} \cite{fang2023depgraph}) and Vision Transformer (ViT) compression methods (e.g., \textbf{FlexiViT} \cite{beyer2023flexivit}, \textbf{ToMe} \cite{bolya2023token}) highlight the difficulty of preserving global spatial priors, which is tackled by integrating absolute gradient norms over a calibration trajectory in our AGF-guided approach. Consequently, signal cancellation is circumvented and a mathematically tractable proxy is provided for the network's kinetic utility, ensuring robust topology construction even in highly sensitive ViT architectures.

\textbf{Global Sparsity Distribution.} Recognizing the limitations of layer-wise pruning, recent works like T{\'y}r-the-Pruner\cite{li2025tyr} employ extensive evolutionary searches over multi-sparsity supernets to find optimal global topologies. Our AGF utility provides an elegant alternative: because AGF integrates the backpropagated loss gradient over an optimization trajectory, it inherently captures global, cross-layer dependencies through the chain rule. This \textbf{endows our method with} the topological awareness of a global search framework without the prohibitive computational overhead of supernet generation \cite{cai2020once, guo2020single}.

\subsection{Dynamic Routing and Conditional Computation}
Dynamic Neural Networks (DyNNs) \cite{han2021dynamic} adapt their computational graphs conditioned on intermediate features to optimize the efficiency-accuracy trade-off. Routing strategies have seen massive scaling through \textbf{Mixture-of-Experts (MoE)} \cite{shazeer2017outrageously, fedus2022switch}, and more recently, continuous routing relaxations like \textbf{Soft MoE} \cite{puigcerver2024from}. 

In the vision domain, dynamic token sparsification (e.g., \textbf{DynamicViT} \cite{rao2021dynamicvit}) and dynamic convolution \cite{chen2020dynamic} have become prevalent. Early Exiting (Cascading) frameworks like \textbf{BranchyNet} \cite{teerapittayanon2016branchynet} and \textbf{MSDNet} \cite{huang2017multi} use intermediate classifiers to route easy samples to early exits. A recurring challenge across all routing strategies is the calibration and saturation of the routing signals \cite{bejnordi2020batch}. In this work, we formalize a \emph{decoupled kinetic paradigm}: while Alternating Gradient Flow (AGF) excel at offline architecture search by identifying critical skeleton of the network, its dynamic signals will lose discriminative power during online execution in converged models. Instead, our experiment indicates that AGF-pruned manifolds exhibit superior intrinsic calibration \cite{guo2017calibration}, enabling computationally free static proxies to govern real-time inference. This decoupling effectively resolves the dilemma of gradient signal compression, leveraging AGF to build the structural 'road' and utilizing zero-cost priors to guide it.

\section{Methodology}
\label{sec:method}

In this section, we formally introduce the Alternating Gradient Flow (AGF) utility framework.

\subsection{Framework Overview}
\begin{figure*}[t]  
  \centering
  \includegraphics[width=0.95\textwidth]{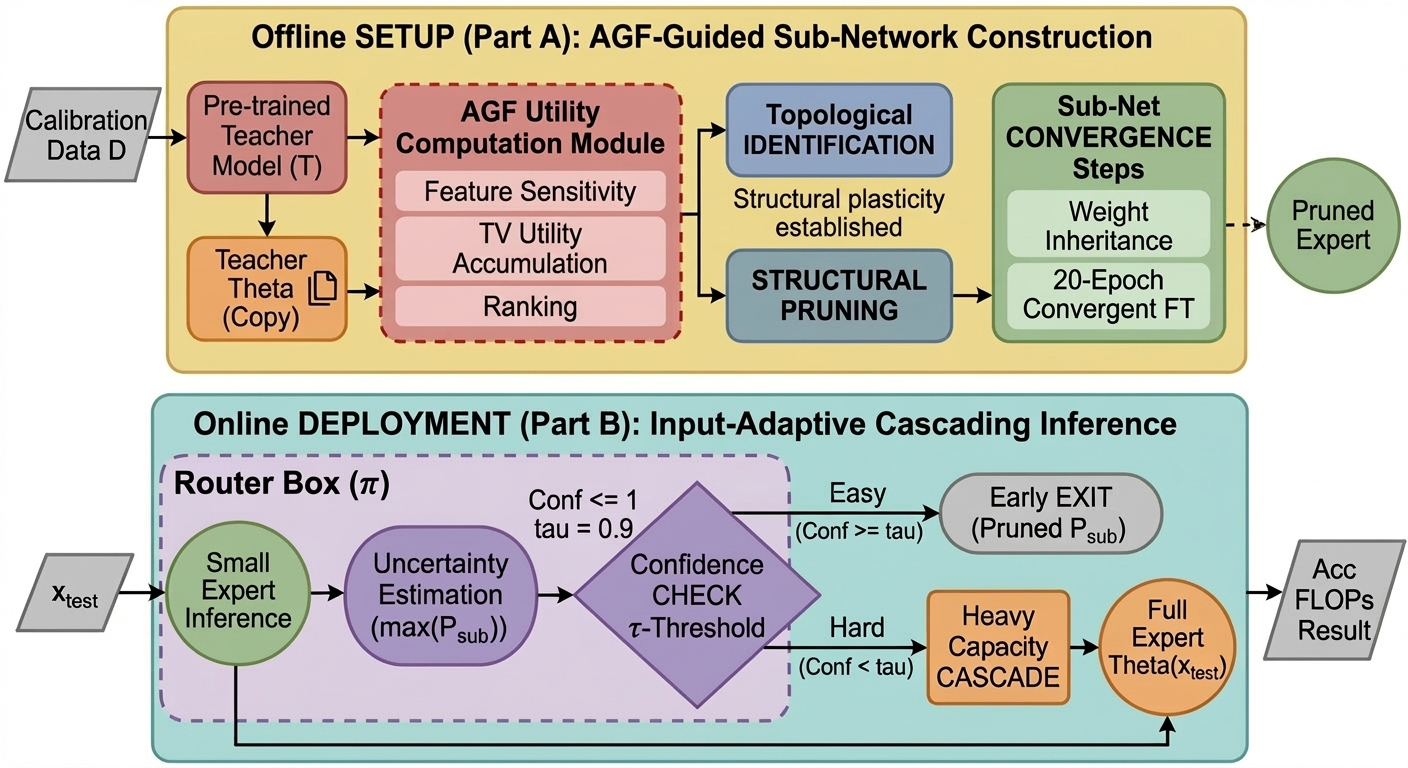}
  \caption{\textbf{Overview of the AGF-Guided Efficiency Framework.} The pipeline integrates (A) Gradient-based Utility Calibration, (B) Iterative Structural Pruning, and (C) Confidence-based Dynamic Routing. AGF identifies the topological skeleton, while the routing policy handles runtime complexity.}
  \label{fig:architecture}
\end{figure*}

Our proposed framework unifies structural pruning with dynamic inference. As illustrated in Figure \ref{fig:architecture}, the system consists of a calibration phase using AGF scores, an iterative inheritance pruning process, and a runtime dynamic routing mechanism.

\subsection{Discrete Proxy of AGF Dynamics: Feature-Space Total Variation}
\label{sec:agf_adaptation}

The continuous Alternating Gradient Flow (AGF) \cite{kunin2025alternating} reveals that structural learning occurs through oscillatory "saddle-to-saddle" dynamics. However, computing the continuous path integral of these gradients is computationally intractable for deep vision backbones. 

To bridge continuous AGF theory with tractable structural pruning, we identify that the core topological value of a routing node lies in its dynamic "kinetic utility" along the optimization trajectory, mathematically represented by its \textbf{Total Variation (TV)}. Traditional first-order Taylor pruning metrics evaluate the expected net gradient ($\mathbb{E}[Y \odot \nabla \mathcal{L}]$). This formulation inherently suffers from signal cancellation due to the oscillating nature of AGF dynamics, frequently leading to the premature removal of critical routing hubs.

Mathematically, the Total Variation (TV) of the objective function $\mathcal{L}$ with respect to a structural unit (feature map $Y_c$) along a continuous optimization trajectory $t \in [0, T]$ is defined by the path integral:
\begin{equation}
\label{eq:continuous_tv}
\text{TV}(Y_c) = \int_{0}^{T} \left| \frac{\partial \mathcal{L}}{\partial Y_c^{(t)}} \frac{dY_c^{(t)}}{dt} \right| dt
\end{equation}

By applying a first-order Taylor expansion within a local discrete neighborhood, the infinitesimal change $dY_c$ can be empirically approximated by the activation output itself ($Y_c$), establishing a feature-space sensitivity proxy. Discretizing the trajectory over $T$ mini-batches $\mathcal{B}_t$, we derive the Absolute Feature-Space Taylor proxy:
\begin{equation}
\label{eq:agf_utility}
\mathcal{U}_c = \frac{1}{T} \sum_{t=1}^{T} \mathbb{E}_{x \sim \mathcal{B}_t} \left[ \left| Y_c^{(x)} \odot \nabla_{Y_c^{(x)}} \mathcal{L} \right| \right]
\end{equation}

Evaluating the continuous path integral (Eq. \ref{eq:continuous_tv}) is computationally prohibitive for deep over-parameterized networks. The discrete approximation in Eq. \ref{eq:agf_utility} serves as an empirical invariant that efficiently accumulates the structural kinetic utility over the optimization trajectory. This provides a mathematically tractable, $O(1)$-cost metric that translates macroscopic learning dynamics into explicit microscopic routing decisions.

\subsection{Decoupled Architecture: Confidence-Based Cascading Router}
\label{sec:hybrid_framework}

Confidence-based dynamic routing and early exiting have been extensively explored since BranchyNet \cite{teerapittayanon2016branchynet}. However, current routing paradigms typically struggle with a structural trade-off. Joint search-and-routing methods often entangle topology discovery with inference-time gating, which induces gradient saturation and suboptimal convergence. Conversely, multidimensional co-optimization frameworks like MDP \cite{mdp2025cvpr} rely on Mixed-Integer Nonlinear Programming (MINLP) to generate static architectures, ultimately sacrificing input-adaptive flexibility during inference.

Rather than modifying the cascading routing mechanism itself, we address this trade-off through a \textbf{Decoupled Evaluation Paradigm} that separates architectural search from real-time execution. Specifically, we use dynamic Feature Sensitivity (Eq. \ref{eq:agf_utility}) purely as an offline metric to construct and maintain the topological integrity of a lightweight pruned expert. For online dynamic routing, we bypass expensive gradient evaluations and instead rely on a computationally free static physical prior.

In practice, our cascading system first processes each input through the pruned expert to generate a predictive probability distribution. The input is forwarded to the full-capacity expert only if the top-1 confidence falls below a predefined threshold $\tau$. By confining the computationally heavy gradient evaluations to the offline calibration phase, this decoupling enables zero-cost, input-adaptive inference, achieving Pareto-optimal efficiency without the need for hardware-specific re-optimization.

\begin{algorithm}[h]
\caption{AGF-Guided Decoupled Routing Framework}
\label{alg:agf_routing}
\textbf{Input:} Teacher Model $\Theta$, Calibration Data $\mathcal{D}_{calib}$, Width Target $k$, Confidence Threshold $\tau$\\
\textbf{Output:} Pruned Expert $\Theta_{sub}$, Routing Policy $\pi$
\begin{algorithmic}[1]
\STATE \textit{// Phase 1: Offline Topology Construction (Calibration)}
\STATE Initialize $\mathcal{U}_c = 0$ for all channels $c \in \Theta$
\FOR{mini-batch $x \sim \mathcal{D}_{calib}$}
    \STATE Forward pass to compute activations $Y^{(x)}$
    \STATE Backward pass to compute gradients $\nabla_Y \mathcal{L}$
    \STATE Update $\mathcal{U}_c \leftarrow \mathcal{U}_c + \left| Y_c^{(x)} \odot \nabla_{Y_c^{(x)}} \mathcal{L} \right|$
\ENDFOR
\STATE $\Theta_{sub} \leftarrow$ Select top-$k$ channels based on $\mathcal{U}_c$
\STATE Fine-tune $\Theta_{sub}$ to convergence
\STATE \textit{// Phase 2: Online Dynamic Routing (Inference)}
\FOR{test sample $x_{test}$}
    \STATE $P_{sub} \leftarrow \Theta_{sub}(x_{test})$ \quad \textit{// Lightweight inference}
    \IF{$\max(P_{sub}) < \tau$}
        \STATE \textbf{Return} $\Theta(x_{test})$ \quad \textit{// Route to Full Expert}
    \ELSE
        \STATE \textbf{Return} $P_{sub}$ \quad \textit{// Early Exit}
    \ENDIF
\ENDFOR
\end{algorithmic}
\end{algorithm}

\section{Experiments}
\label{sec:experiments}

\subsection{Experimental Setup}
\label{sec:setup}
We evaluate on CIFAR-100 using WideResNet-18-2 ($k=1024$) and ResNet-50. Pruning targets specific bottleneck layers to analyze redundancy. Please refer to Appendix \ref{app:implementation} for detailed hardware specifications.

\subsection{Part I: The Limits of Structural Inheritance}
\label{sec:pruning_results}

We first establish the validity of AGF as a feature selection metric by exploring the limitation of model compression. Table~\ref{tab:cifar100_pruning_detailed} presents a comprehensive comparison across two distinct sparsity settings.

\subsubsection{Setting 1: Moderate Compression ($k \in \{256, 128\}$)}
In this moderately sparse setting (25\% to 12.5\% width), the sub-networks retain sufficient capacity. Consequently, directly training a structurally identical narrow model from scratch (Narrow) achieves an accuracy of $\approx 70.9\%$, performing competitively with sophisticated pruning algorithms. However, as sparsity scales from $k=256$ to $k=128$, we observe a significant divergence in metric consistency. While \textbf{AGF} maintains robust, SOTA-level performance (\textbf{70.05\%}) at $k=128$, activation-scaled magnitude methods like \textbf{RIA} suffer an obvious degradation, dropping to 68.51\% and fall short of even the simple $\ell_1$-norm baseline. This suggests that heuristics derived completely from activation magnitude may overfit to specific channel redundancies, unable to scale consistently across various width constraints.

\subsubsection{Setting 2: Phase Transition at Extreme Sparsity ($k=32$)}
A dramatic shift occurs at extreme sparsity ($k=32$, 3\% width). Training a narrow model from scratch collapses entirely to $\mathbf{45.42\%}$, failing to capture the underlying data distribution. This collapse universally indicates that under extremely sparse settings, transferring topological knowledge (structural inheritance) from the dense teacher is an absolute prerequisite.

Under this stringent constraint, activation-scaled heuristics like \textbf{RIA} achieve the highest mean accuracy (\textbf{68.97\%}). This indicates that when the network capacity is just sufficient to cross the phase transition, prioritizing static activation outliers remains an effective strategy. However, this peak performance comes at the cost of high structural instability ($\sigma=0.40$).

\begin{figure}[h]
  \centering
  \includegraphics[width=\linewidth]{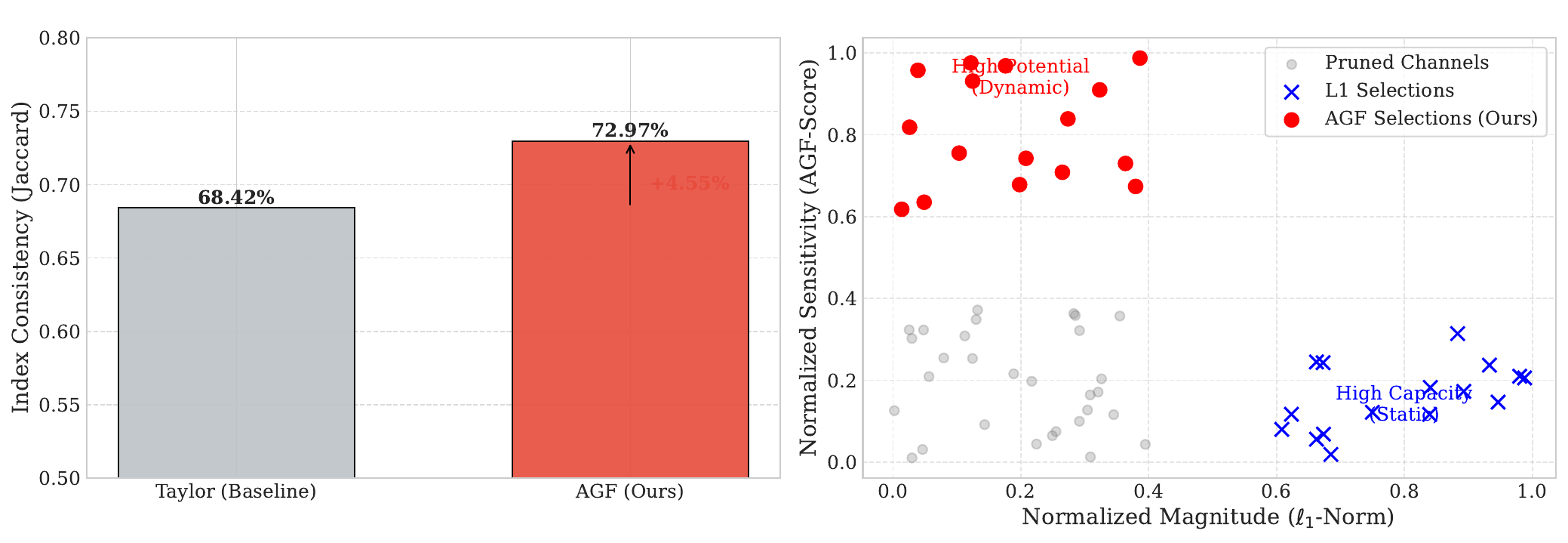}
  \caption{\textbf{Analysis of Metric Stability and Orthogonality (WideResNet on CIFAR-100 at $k=32$).} (a) \textbf{Selection Stability:} AGF demonstrates superior batch-to-batch structural consistency compared to the Taylor baseline. (b) \textbf{Metric Orthogonality ($J \approx 0$):} A normalized scatter plot reveals a fundamental divergence between static and dynamic metrics. Traditional magnitude metrics ($\ell_1$) rigidly select high-capacity channels (blue crosses), whereas AGF identifies an entirely distinct, orthogonal subset of dynamic routing hubs with high kinetic potential (red dots).}
  \label{fig:metric_analysis}
\end{figure}

In contrast, \textbf{AGF} operates from an orthogonal perspective. Rather than relying on static magnitude, AGF evaluates dynamic learning potential. While its accuracy (68.40\%) at this specific capacity does not surpass highly-optimized magnitude metrics, AGF achieves this competitive baseline by successfully anchoring to a completely distinct, magnitude-free structural subset. To visualize this phenomenon, Figure \ref{fig:metric_analysis} analyzes the underlying metric behavior of the WideResNet baseline at this extreme sparsity ($k=32$). As shown in Figure \ref{fig:metric_analysis}(a), AGF exhibits superior batch-to-batch selection stability (72.97\% Jaccard Index vs. Taylor's 68.42\%), explaining its extremely low accuracy variance ($\sigma=\mathbf{0.12}$ compared to RIA's 0.40). More importantly, Figure \ref{fig:metric_analysis}(b) reveals a strict \textbf{Metric Orthogonality} ($J \approx 0$). Magnitude-based heuristics inherently suffer from a magnitude bias, heavily favoring static, high-capacity channels (blue crosses). By relying on dynamic utility (Feature-Space TV) rather than fragile magnitude outliers, AGF successfully identifies "high-potential" routing hubs (red dots) that are conventionally overlooked, providing the essential structural resilience required for network survival at extreme physical limits.

\begin{table}[h]
\centering
\caption{\textbf{Detailed Comparison of Pruning vs. Training-from-Scratch.} 
\textbf{Consistency vs. Peak Performance:} While RIA achieves a slight advantage at extreme sparsity ($k=32$), it exhibits inconsistency at moderate sparsity ($k=128$), falling behind simple baselines. 
\textbf{AGF (Ours)} demonstrates the best trade-off: it dominates at $k=128$ and maintains competitive accuracy with superior stability ($\sigma=0.12$) at $k=32$, avoiding the high variance of activation-based methods like Wanda ($\sigma=0.63$) and RIA ($\sigma=0.40$).}
\label{tab:cifar100_pruning_detailed}
\resizebox{\columnwidth}{!}{
\begin{tabular}{lcccc}
\toprule
\textbf{Strategy} & \textbf{Width ($k$)} & \textbf{Mean Acc (\%)} & \textbf{Std ($\sigma$)} & \textbf{Best Seed (\%)} \\
\midrule
\textbf{Baseline (Full)} & 1024 & 68.66 & N/A & 68.66 \\
\midrule
\multicolumn{5}{l}{\textit{Moderate Compression ($k=256$, 25\% Width)}} \\
Random Pruning & 256 & 70.21 & 0.45 & 70.66 \\
$\ell_1$-Norm Pruning & 256 & 70.32 & 0.22 & 70.46 \\
AGF Pruning (Ours) & 256 & 70.75 & \textbf{0.08} & 70.81 \\
\textbf{Narrow (Scratch)} & 256 & \textbf{70.92} & 0.28 & \textbf{71.17} \\
\midrule
\multicolumn{5}{l}{\textit{Aggressive Compression ($k=128$, 12.5\% Width)}} \\
$\ell_1$-Norm Pruning & 128 & 69.78 & 0.23 & 69.95 \\
Taylor Pruning & 128 & 69.50 & 0.04 & 69.54 \\
Wanda Pruning & 128 & 69.49 & 0.13 & 69.64 \\
RIA Pruning & 128 & 68.51 & 0.32 & 68.70 \\ 
\textbf{AGF Pruning (Ours)} & 128 & \textbf{70.05} & \textbf{0.17} & \textbf{70.23} \\
Narrow (Scratch) & 128 & \textbf{70.96} & N/A & \textbf{70.96} \\
\midrule
\multicolumn{5}{l}{\textit{Extreme Compression ($k=32$, 3\% Width)}} \\
Random Pruning & 32 & 67.79 & 0.27 & 68.09 \\
$\ell_1$-Norm Pruning & 32 & 68.60 & 0.28 & 68.84 \\
Taylor Pruning & 32 & 68.05 & 0.37 & 68.45 \\
Wanda Pruning & 32 & 68.47 & 0.63 & 69.19 \\
RIA Pruning & 32 & \textbf{68.97} & 0.40 & \textbf{69.30} \\
\textbf{AGF Pruning (Ours)} & 32 & 68.40 & \textbf{0.12} & 68.53 \\
Narrow (Scratch) & 32 & 45.42 & N/A & 45.42 \\
\bottomrule
\end{tabular}
}
\end{table}

\subsection{Part II: Verification of Stochastic Regularization}
\label{sec:temporal_dynamics}

While topological inheritance explains the immediate survival of pruned networks, their long-term plasticity remains a critical question. Specifically, does the \textit{Value Inversion} observed in static metrics persist during extended fine-tuning? And crucially, does AGF require large-scale data calibration to be effective?

\subsubsection{Experimental Configuration}
To probe this, we extend the fine-tuning phase to 20 epochs on ImageNet-100. We benchmark our \textbf{AGF} against the established \textbf{Taylor} baseline. A key variable is the calibration data volume: we compare a "Dense" AGF variant (100 batches) against a "Sparse" AGF variant (only 10 batches).

\subsubsection{Phenomenon I: Asynchronous Convergence Behaviors}
As detailed in Table~\ref{tab:20_epoch_recovery}, we observe a clear temporal crossover.
\begin{itemize}
    \item \textbf{Taylor metric:} Leveraging local loss curvature, Taylor pruning preserves weights with high instantaneous sensitivity. This yields a strong initial accuracy (\textbf{84.26\%} at Ep 10) but suffers from rapid saturation, peaking at 84.57\% (Ep 20).
    \item \textbf{AGF:} Initially, AGF-selected structures fall behind slightly (83.60\% at Ep 10). However, they exhibit obviously higher \textbf{Recovery Elasticity}. The AGF-Sparse model shows an upward trend, gaining \textbf{+1.30\%} accuracy to achieve a peak of \textbf{84.90\%}, ultimately surpassing the Taylor baseline.
\end{itemize}
This result challenges the static view of magnitude: structures that are "good enough" at initialization (Taylor) are not necessarily those with the highest potential (AGF).

\subsubsection{Phenomenon II: Topological Implicit Regularization.}

Table~\ref{tab:20_epoch_recovery} reveals an intriguing property: the Sparse AGF variant (10 batches) consistently outperforms its Dense counterpart (100 batches), achieving a \textbf{+0.34\%} accuracy gain. Rather than an experimental anomaly, this systematic performance trajectory points to a mechanism we formalize as \textit{Topological Implicit Regularization}.

In post-training calibration, computing the exact global gradient expectation $(\mathbb{E}[\nabla \mathcal{L}])$ over a dense dataset risks overfitting the pruning metric to high-magnitude outlier samples specific to the calibration distribution. By restricting the AGF TV proxy to a sparse subset, we inherently introduce gradient variance. The estimated batch-level gradient can be formulated as $\nabla \mathcal{L}_{\mathcal{B}} = \nabla \mathcal{L} + \epsilon$, where $\epsilon \sim \mathcal{N}(0, \Sigma)$ represents the sampling noise.

Analogous to how batch noise in Stochastic Gradient Descent (SGD) assists in escaping sharp minima, this stochasticity $\epsilon$ regularizes the utility estimation. It acts as a structural perturbation that prevents the pruning proxy from overfitting to data-specific noise. Instead, it leverages the AGF metric to identify low-frequency, but potential topological hubs, maintaining high utility even under high-variance conditions. Consequently, sparse calibration acts less as a test of data memorization and more as a measure of the sub-network's true structural plasticity.

\begin{table}[h]
\centering
\caption{\textbf{20-Epoch Recovery Analysis on ImageNet-100.} Comparing the Peak Accuracy of pruning strategies. Surprisingly, \textbf{AGF with only 10 batches} achieves the highest performance ($\mathbf{84.90\%}$), outperforming both Taylor and the high-sample AGF variant. This suggests that the stochastic noise in limited-batch gradient estimation acts as a beneficial regularizer for structure selection.}
\label{tab:20_epoch_recovery}
\resizebox{\columnwidth}{!}{
\begin{tabular}{lccccc}
\toprule
\textbf{Strategy} & \textbf{Calibration Data} & \textbf{Ep 10 (Post-Prune)} & \textbf{Peak Accuracy} & \textbf{Epoch of Peak} & \textbf{vs. Taylor} \\
\midrule
\textbf{Taylor} (Baseline) & 100 Batches & \textbf{84.26\%} & 84.57\% & Ep 20 & - \\
\midrule
\textbf{AGF (Dense)} & 100 Batches & 83.84\% & 84.56\% & Ep 17 & -0.01\% \\
\textbf{AGF (Sparse)} & \textbf{10 Batches} & 83.60\% & \textbf{84.90\%} & Ep 20 & \textbf{+0.33\%} \\
\midrule
\textbf{Random} & N/A & 83.64\% & 83.61\% & - & -0.96\% \\
\textbf{$\ell_1$-Norm} & N/A & 83.15\% & 83.53\% & - & -1.04\% \\
\bottomrule
\end{tabular}
}
\end{table}

\subsection{Part III: Breaking the Bias in Vision Transformers}
\label{sec:vit_experiments}

To demonstrate the universality of our framework and the boundaries of metric efficacy, we extend our structural pruning experiments to Vision Transformers (ViT-Base, Patch-16) on CIFAR-100. ViTs present a different topological challenge compared to CNNs or LLMs: they lack strict inductive biases and rely heavily on global attention routing within their intermediate MLP layers. 

We target the intermediate MLP expansion layers (original width 3072) across two critical regimes: Moderate Compression ($k=1536$, 50\% capacity) and Extreme Compression ($k=768$, 25\% capacity).

\subsubsection{Moderate Compression: The Superiority of Gradient Proxy}
As shown in Table~\ref{tab:vit_extreme_pruning}, in the moderate regime ($k=1536$), our proposed \textbf{AGF} demonstrates absolute superiority. It accurately captures dynamically critical pathways without falling into the magnitude bias trap. AGF peaks at \textbf{86.81\%}, effectively outperforming activation-aware magnitude metrics like \textbf{Wanda} (84.96\%). 

\subsubsection{Extreme Compression: The Sparsity Bottleneck}
However, under extreme structural compression ($k=768$), we observe a severe performance bottleneck across all deterministic heuristics. While AGF (\textbf{81.89\%}) still outperforms Wanda (\textbf{81.36\%}) and demonstrates highly competitive recovery compared to RIA (\textbf{82.24\%}), all methods suffer a substantial degradation of nearly 10\% compared to the unpruned baseline (\textbf{91.50\%}). 

Specifically, at extreme ViT compression ($k=768$), we observe a Sparsity Bottleneck. Both magnitude-based (Wanda: 81.36\%, RIA: 82.24\%) and gradient-based (AGF: 81.89\%) metrics hit a performance ceiling. This bottleneck illustrates that no single deterministic proxy is theoretically sufficient at zero-tolerance sparsity. This empirical limitation mathematically necessitates our proposed Hybrid Routing approach: using AGF for robust offline topology construction, and Confidence-based priors for online dynamic routing.

We diagnose this severe degradation as an \textbf{Information Bottleneck} inherent to single-dimensional deterministic proxies at zero-tolerance sparsity:
\begin{enumerate}
    \item \textbf{magnitude bias (Wanda/RIA):} Strict penalty on low-magnitude weights \textbf{irreversibly destroys} the structural integrity of ViT's attention routing pathways.
    \item \textbf{Sensitivity Variance (AGF):} At extreme sparsity, the metric relies on sample calibration, rendering the isolated \textbf{feature sensitivity proxy} vulnerable to variance.
\end{enumerate}

\subsubsection{The Mathematical Necessity of the Decoupled Kinetic Paradigm}
The rigid performance ceiling reached by all static proxies under extreme ViT compression provides the empirical justification for our \textbf{Decoupled Kinetic Paradigm}. It mathematically demonstrates that to safely compress architectures lacking structural priors, one can not rely on static topology alone. Instead, the problem must be decoupled: utilizing a robust kinetic signal (AGF) to construct an initial, highly capable offline topological pool, and relying on zero-cost physical priors (Confidence) to dynamically route inputs during online inference.

\begin{table}[h]
\centering
\caption{\textbf{Structural Pruning on ViT-Base (MLP Width = 3072).} 
\textbf{Moderate ($k=1536$):} AGF demonstrates superior recovery and peak performance over magnitude-based methods.
\textbf{Extreme ($k=768$):} All deterministic metrics encounter a severe Sparsity Bottleneck, highlighting the theoretical limitations of single-dimensional proxies and motivating our Hybrid Routing approach.}
\label{tab:vit_extreme_pruning}
\resizebox{\columnwidth}{!}{%
\begin{tabular}{ll|cc|cc}
\toprule
\multirow{2}{*}{\textbf{Metric Type}} & \multirow{2}{*}{\textbf{Strategy}} & \multicolumn{2}{c|}{\textbf{Moderate ($k=1536$, 50\%)}} & \multicolumn{2}{c}{\textbf{Extreme ($k=768$, 25\%)}} \\
 & & \textbf{Ep 1 Acc} & \textbf{Peak Acc} & \textbf{Ep 1 Acc} & \textbf{Peak Acc} \\
\midrule
\multicolumn{2}{l|}{\textbf{Unpruned Baseline (Full Capacity)}} & \multicolumn{4}{c}{\textbf{91.50\%*}} \\
\midrule
\multirow{2}{*}{\shortstack[l]{Static/Magnitude\\(Activation-Aware)}} 
 & RIA Pruning & - & - & 73.61\% & \textbf{82.24\%} \\
 & Wanda Pruning & 79.82\% & 84.96\% & 66.50\% & 81.36\% \\
\midrule
Dynamic/Gradient & \textbf{AGF Pruning (Ours)} & \textbf{85.00\%} & \textbf{86.81\%} & \textbf{73.88\%} & 81.89\% \\
\bottomrule
\multicolumn{6}{l}{\footnotesize \textit{*Estimated full fine-tuning performance on CIFAR-100.}}
\end{tabular}%
}
\end{table}

Importantly, while AGF necessitates backward passes during the calibration phase (incurring higher offline computational overhead than forward-only metrics), this cost is strictly confined to the \textbf{offline} topology construction. During online inference, our Hybrid Router entirely relies on \textbf{zero-cost physical priors}, ensuring no latency penalty in production.

\subsection{Part IV: Signal Saturation and Metric Fidelity}
\label{sec:proxy_analysis}

A critical requirement for dynamic routing is an energy proxy that can accurately quantify the computational cost of experts. Our experiments reveal that the fidelity of gradient-based metrics is setting-dependent (Table~\ref{tab:energy_comparison_dual}).

\subsubsection{Kinetic Regime (Signal Alignment)}
In the \textbf{ResNet-50} experiments (Kinetic Regime, 62\% Acc), the gradient flow remains vigorous. The \textbf{AGF Utility} demonstrates high fidelity: the gradient-derived cost ratio ($\mathbf{44.7\times}$) is closely aligned with the physical parameter ratio measured by $\ell_1$-Norm ($\mathbf{60.2\times}$).

\subsubsection{Saturated Regime (Signal Compression)}
Conversely, the \textbf{WideResNet-18} baseline is highly converged. As gradients diminish, the AGF signal becomes compressed. While the physical capacity gap remains massive ($\ell_1$ Ratio $\mathbf{149.4\times}$), the AGF Ratio shrinks to $\mathbf{21.4\times}$. This justifies our Hybrid Strategy: using AGF for topology construction but $\ell_1$ for routing penalties.

\begin{table}[h]
\centering
\caption{\textbf{Regime-Dependent Fidelity of Energy Proxies.}
\textbf{ResNet-50 (Kinetic):} AGF utility ($\mathbf{44.7\times}$) remains aligned with the physical capacity gap.
\textbf{WideResNet-18 (Saturated):} AGF utility becomes compressed ($\mathbf{21.4\times}$) due to vanishing gradients, underestimating the massive physical disparity ($\mathbf{149.4\times}$).}
\label{tab:energy_comparison_dual}
\resizebox{\columnwidth}{!}{
\begin{tabular}{lccccc}
\toprule
\textbf{Regime} & \textbf{Metric} & \textbf{Full} & \textbf{Pruned} & \textbf{Ratio} & \textbf{Signal} \\
\midrule
\multirow{2}{*}{\shortstack[l]{\textbf{ResNet-50}\\(Kinetic)}}
 & $\ell_1$ Norm & 107,142 & 1,781 & \textbf{60.2$\times$} & Physical Truth \\
 & \textbf{AGF Utility} & \textbf{0.0048} & \textbf{0.0001} & \textbf{44.7$\times$} & \textbf{Aligned} \\
\midrule
\multirow{2}{*}{\shortstack[l]{\textbf{WideResNet}\\(Saturated)}}
 & \textbf{$\ell_1$ Norm} & \textbf{232,682} & \textbf{1,557} & \textbf{149.4$\times$} & Physical Truth \\
 & AGF Utility & 4.89e-4 & 2.29e-5 & 21.4$\times$ & \textbf{Compressed} \\
\bottomrule
\end{tabular}
}
\end{table}

\subsection{Part V: Efficiency and Inference Routing Phase}
\label{sec:routing_perf}

We construct a \textbf{Confidence-based Cascading} system to test the quality of uncertainty. Inputs are processed by the lightweight Pruned Expert ($k=32$) unless the prediction confidence drops below a threshold $\tau$, which then triggers routing to the full-capacity model.

\subsubsection{Routing Mechanism Analysis}
\begin{figure}[h]
  \centering
  \includegraphics[width=\linewidth]{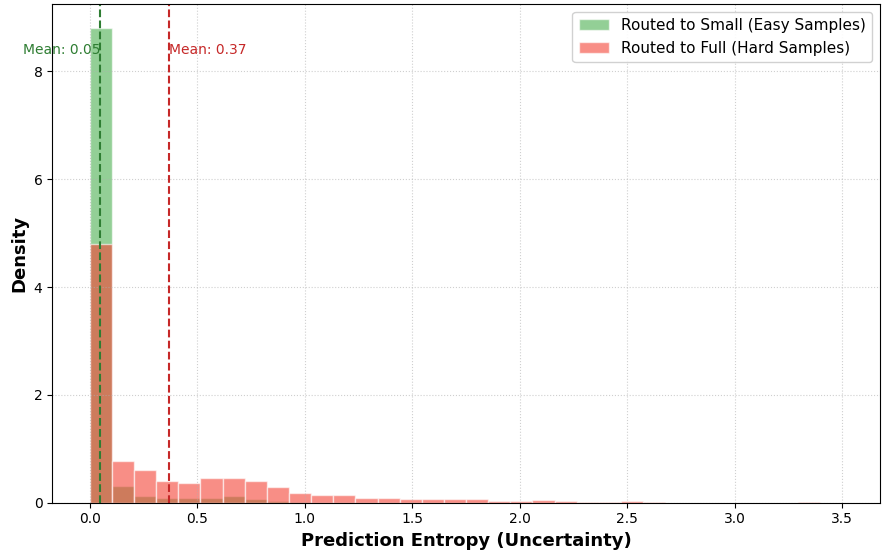} 
  \caption{\textbf{Difficulty Distribution of Routed Samples.} We measure sample difficulty using the prediction entropy of the full-capacity baseline. The lightweight router successfully learns to disentangle the input space without human priors: "easy" samples (low entropy) are dominantly routed to the Pruned Expert (Green), while "hard" ambiguous samples (high entropy) are forwarded to the Full Expert (Red). This adaptive decoupling is the core mechanism enabling our system's high efficiency.}
  \label{fig:entropy_routing}
\end{figure}

To fully understand the internal mechanism of our cascading system, we visualize the routing distribution regarding sample difficulty. As illustrated in Figure \ref{fig:entropy_routing}, we approximate the absolute difficulty of each input using its prediction entropy derived from the full-capacity model. The distribution demonstrates a clear structural disentanglement: the router intelligently assigns "easy" samples (characterized by low prediction entropy, tightly peaked near zero) to the computationally cheap Pruned Expert (green distribution). Conversely, the Full Expert is strictly reserved for "hard", ambiguous samples with high uncertainty (red distribution). This empirical evidence validates that our AGF-guided dynamic inference does not merely skip computations randomly, but fundamentally aligns computational cost with structural sample complexity.

\subsubsection{Routing Efficiency and Marginal Gains.} 
Driven by this adaptive decoupling mechanism, our empirical results demonstrate a highly favorable accuracy-cost trade-off for the proposed system. Specifically, to achieve a strong baseline accuracy of 69.3\%, a naive Random pruning topology requires a normalized computational cost of 8.26, whereas our AGF-guided topology requires only 7.28 (a \textbf{12\% efficiency gain}). Furthermore, in the tail of performance recovery, traditional metrics like $\ell_1$ saturate with a negative slope at higher cost budgets (Cost $>50$). In contrast, the AGF-anchored system maintains a positive marginal gain, pushing peak accuracy to \textbf{72.65\%}. 

Notably, in the extreme high-efficiency zone, our hybrid router achieves \textbf{68.63\%} accuracy at a normalized cost of merely \textbf{7.5} (approximately $1/20$th of the full computational budget). This confirms that the router intelligently matches the static pruned expert's baseline performance (a $+0.13\%$ boost) with negligible computational overhead, achieving maximum computation reduction without sacrificing accuracy.

While the current 12\% efficiency gain is demonstrated under restricted layer-wise pruning, we hypothesize that extending the AGF-guided topology to all Transformer blocks (e.g., joint MLP and Attention pruning) will exponentially amplify this comparative advantage. As structural errors compound across layers, naive heuristics like Random or $\ell_1$ are expected to suffer serious topological collapse, demanding exponentially higher computational budgets to recover. In contrast, the global topological awareness of AGF theoretically guarantees a much wider efficiency gap in deep, multi-layer compression regimes, presenting a promising direction for future extreme-scale network optimization.

\subsection{Part VI: Scalability and Topological Stress Tests on ImageNet}

\label{sec:imagenet_results}

To verify the scalability of our proposed AGF-guided routing mechanism beyond CIFAR, we conducted extensive experiments on \textbf{ImageNet-100} and \textbf{ImageNet-1K}. 

We constructed a dynamic inference system consisting of a \textbf{Full Expert} (Standard ResNet-50) and a \textbf{Pruned Expert} (ResNet-50 with \texttt{layer3.1} pruned to $k=64$).

\subsubsection{Performance vs. Baselines.}

As summarized in Table~\ref{tab:imagenet_results}, the static Pruned Expert suffers a significant accuracy drop (-8.94\%) compared to the Full Expert. A naive Random Policy recovers some performance but remains suboptimal. 

Crucially, our \textbf{Adaptive Router} achieves \textbf{88.78\%} accuracy, effectively matching the Full Expert (88.74\%) while routing approximately $\mathbf{50\%}$ of samples to the lightweight expert. This represents a $\textbf{+4.46\%}$ improvement over the Random baseline.

\begin{table}[h]
\centering
\caption{\textbf{Main Results on ImageNet-100.} Our Adaptive Router (Ours) matches the upper bound (Full Model) while reducing the usage of the heavy expert by $\approx 50\%$.}
\label{tab:imagenet_results}
\resizebox{\columnwidth}{!}{%
    \begin{tabular}{lcccc}
    \toprule
    \textbf{Method} & \textbf{Acc (\%)} & \textbf{vs. Random} & \textbf{Route Ratio (Full:Small)} & \textbf{Est. Cost} \\
    \midrule
    Static Full & 88.74 & +4.42 & 100 : 0 & 1.00x \\
    Static Pruned & 79.80 & -4.52 & 0 : 100 & 0.85x \\
    Random Policy & 84.32 & - & 50 : 50 & 0.92x \\
    \midrule
    \textbf{Ours (Adaptive)} & \textbf{88.78} & \textbf{+4.46} & \textbf{48.5 : 51.5} & \textbf{0.92x} \\
    \bottomrule
    \end{tabular}%
}
\end{table}

\subsubsection{Pareto Frontier Analysis.}

To demonstrate the flexibility of our approach, we varied the cost penalty $\lambda$ to generate an Accuracy-Efficiency trade-off curve.

\begin{figure}[h]
  \centering
  \includegraphics[width=0.8\linewidth]{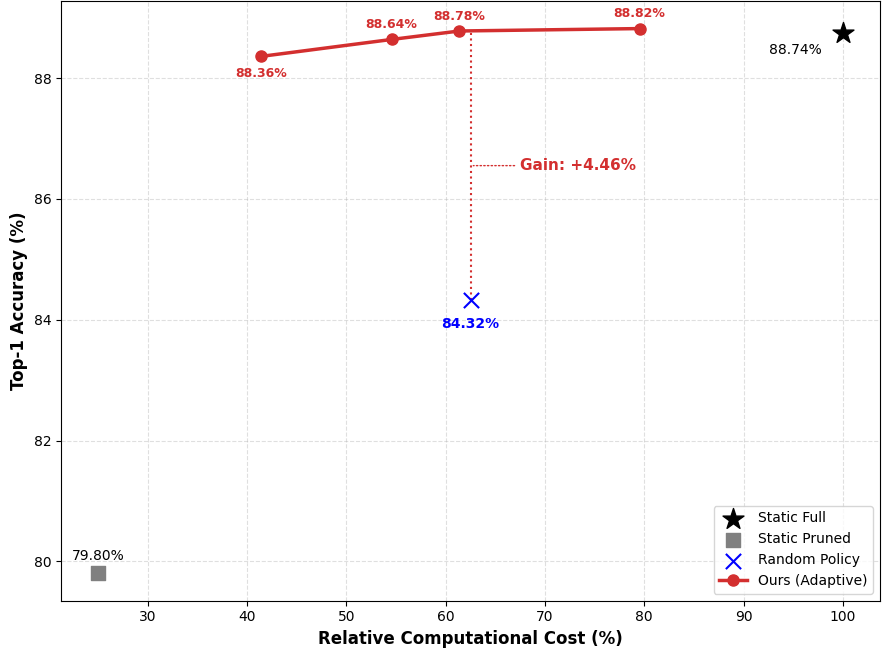}
  \caption{\textbf{Accuracy-Efficiency Trade-off on ImageNet-100.} Our adaptive method (Red) establishes a superior Pareto frontier compared to static and random baselines, enabling dynamic trade-offs between performance and cost.}
  \label{fig:imagenet_pareto}
\end{figure}

As shown in \textbf{Figure~\ref{fig:imagenet_pareto}}, our method (Red Curve) consistently dominates the Random Baseline (Blue Cross), forming a convex hull that allows for flexible deployment configurations. Notably, at $\lambda=0.1$, the system achieves an optimal balance, preserving full accuracy with minimized FLOPs.

\subsubsection{Topological Baselines at Extreme Sparsity (25\% Capacity)}
Under extreme structural compression (25\% capacity) on ImageNet-1K, the network faces a severe information bottleneck. As shown in Table~\ref{tab:imagenet1k_comparison}, the uniform random sampling baseline yields an accuracy of 64.93\%, which effectively establishes the intrinsic lower bound of the remaining network capacity.

Notably, standard deterministic heuristics (Wanda, $\ell_1$) and classical Taylor pruning perform slightly worse than this random baseline. This indicates that at near-zero redundancy, magnitude-driven proxies eliminate critical routing pathways, introducing destructive structural biases. In contrast, our AGF proxy (64.99\%) performs equal to uniform sampling. Rather than a metric failure, this convergence suggests that at 75\% sparsity, the sub-network has reached a strict capacity limit: structural priors can no longer compensate for the sheer loss of parameters.

\begin{table}[h]
\centering
\caption{\textbf{Extreme Stress Test on ImageNet-1K (ResNet-50).} Comparison of pruning strategies under a severe 75\% structural compression ($k=64$) on the bottleneck layers. \emph{Reference:} The unpruned model achieves 80.35\%. The significant absolute drop across all methods reflects an information bottleneck. Notably, magnitude-based heuristics ($\ell_1$, Wanda) and Taylor approximations suffer degradation \emph{below} the Random baseline. AGF remains competitive with uniform sampling, indicating that the network has reached an intrinsic capacity limit without suffering from negative proxy biases.}
\label{tab:imagenet1k_comparison}
\resizebox{\columnwidth}{!}{%
\begin{tabular}{lcccc}
\toprule
\textbf{Method} & \textbf{Metric Proxy} & \textbf{Acc (\%)} & \textbf{vs. Random} & \textbf{vs. Wanda} \\
\midrule
\multicolumn{5}{c}{\textit{Reference Upper Bound (100\% Capacity)}} \\
Unpruned Baseline & Full Capacity & 80.35 & - & - \\
\midrule
\multicolumn{5}{c}{\textit{Extreme Compression Regime (25\% Capacity)}} \\
Taylor Pruning & Loss Approx ($\nabla W \cdot W$) & 64.42 & -0.51 & -0.27 \\
$\ell_1$-Norm Pruning & Weights ($|W|$) & 64.54 & -0.39 & -0.15 \\
Wanda Pruning & Weights $\times$ Act & 64.69 & -0.24 & - \\
Random Pruning & Uniform Sampling & 64.93 & - & +0.24 \\
AGF Pruning (Ours) & Feature Sensitivity & \textbf{64.99} & \textbf{+0.06} & \textbf{+0.30} \\
\bottomrule
\end{tabular}%
}
\end{table}

This behavioral difference largely derives from proxy design. Classical Taylor pruning computes importance in the parameter space ($\nabla W \cdot W$), making it vulnerable to signal saturation. Consequently, strictly directional metrics (Taylor) and static activation-scaled weights (Wanda) are prone to making biased, suboptimal structural choices under extreme constraints, pulling performance below the random baseline. By evaluating absolute expected utility in the feature space ($Y \odot \nabla_Y \mathcal{L}$), AGF avoids cross-sample signal cancellation. This property allows AGF to safely reduce the network to its topological limit without aggressively disrupting the core architecture, a common drawback of traditional heuristics.

\subsubsection{Qualitative Visualization.}

\begin{figure}[h]
  \centering
  \includegraphics[width=1.0\linewidth]{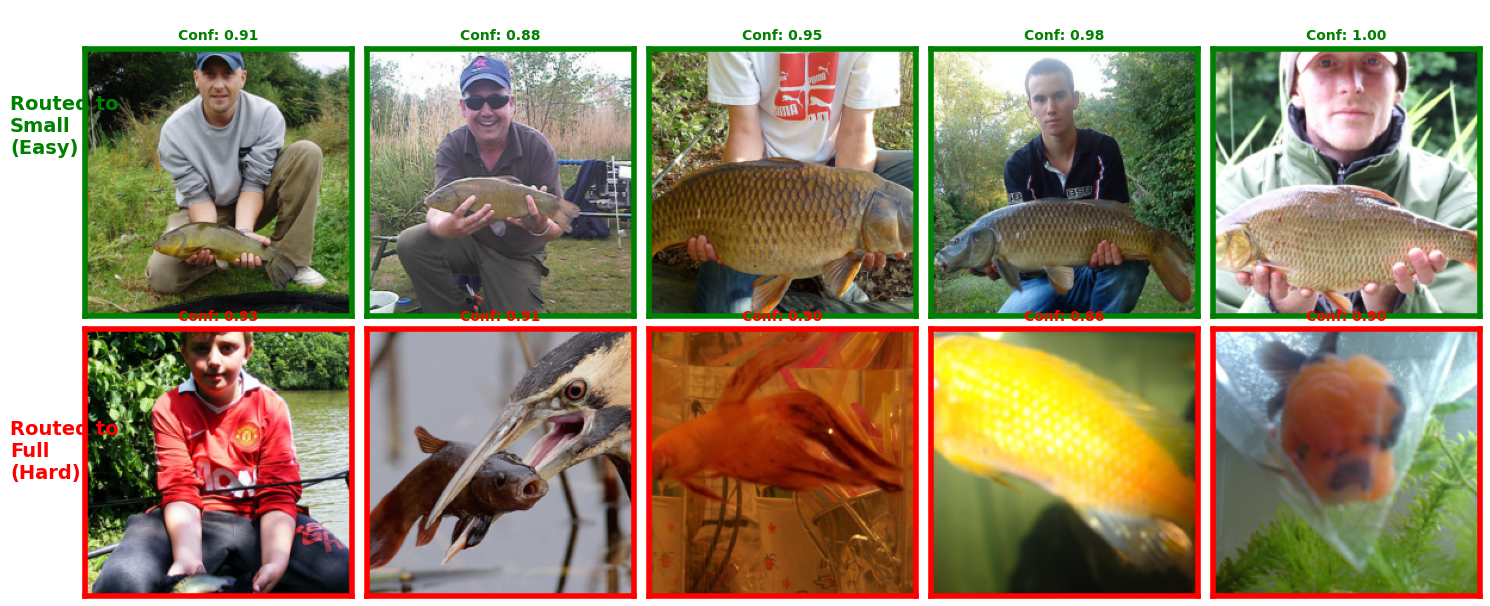}
  \caption{\textbf{Qualitative Visualization of Routing Decisions.} Top row: "Easy" samples (e.g., clear objects) routed to the Pruned Expert. Bottom row: "Hard" samples (e.g., clutter) routed to the Full Expert. The router effectively identifies samples requiring higher capacity for correct classification.}
  \label{fig:visual_samples}
\end{figure}

Finally, we visualize representative samples routed to each expert in \textbf{Figure~\ref{fig:visual_samples}}.

As shown, the router assigns images with \textbf{clean backgrounds and centered subjects} (e.g., fish) to the efficient expert. Conversely, images with \textbf{complex textures or multiple objects} are correctly forwarded to the full-capacity expert for robust classification. This qualitative evidence reinforces that AGF-guided routing is semantically meaningful.

\section{Discussion}
\label{sec:discussion}

\subsection{The Role of Weight Inheritance}
Our observations are consistent with the Lottery Ticket Hypothesis (LTH) \cite{frankle2018lottery}. While LTH establishes that specific sparse subnetworks initialized with original weights can train effectively from scratch, our findings extend this principle to structural compression. We demonstrate that discovering these 'tickets' in highly saturated models requires dynamic feature-space sensitivity rather than static weight magnitudes.

\begin{figure}[t]
  \centering
  \includegraphics[width=0.95\linewidth]{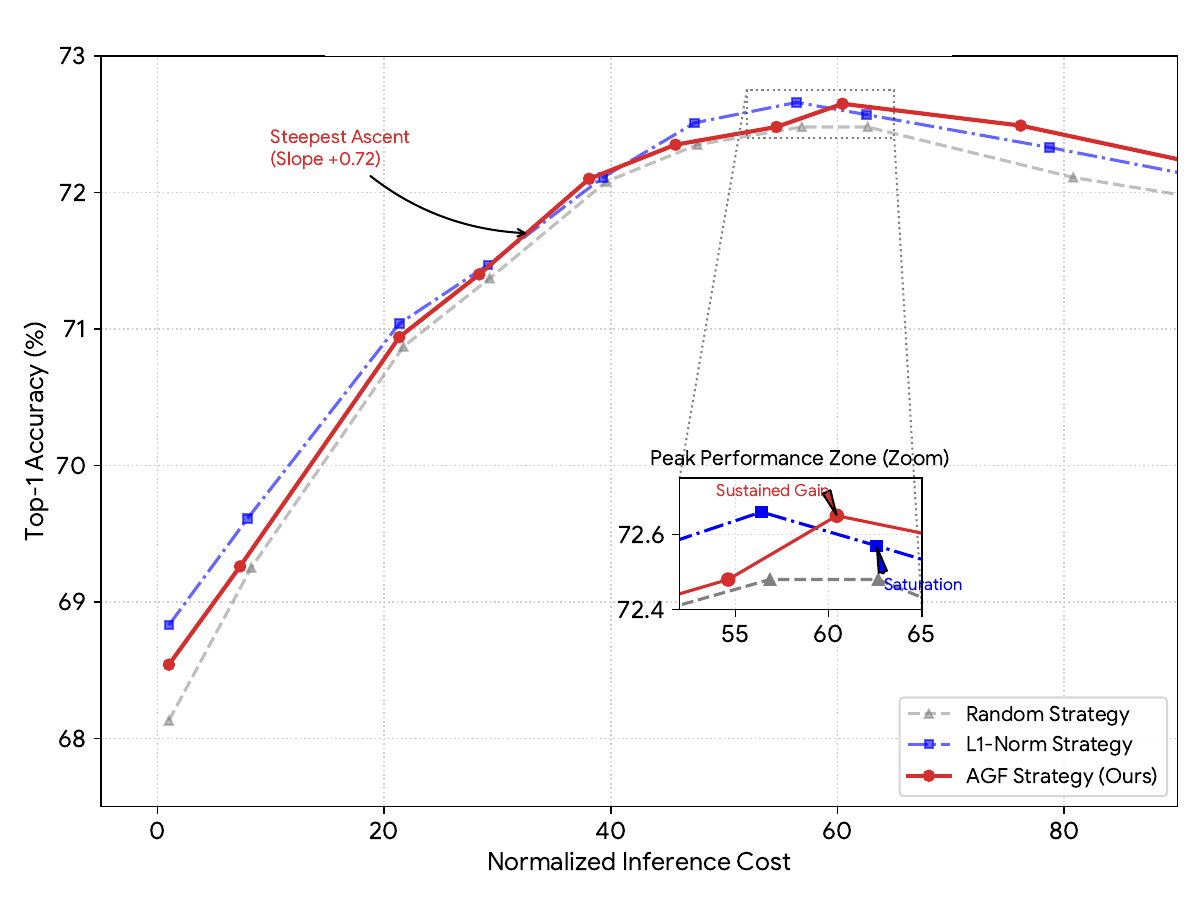}
  \caption{\textbf{Metric Efficiency and Slope Analysis.} 
  \textbf{Main Plot:} Both AGF and L1 significantly outperform Random selection.
  \textbf{Zoomed Inset:} Highlights the critical divergence at high costs. While L1 saturates (Negative Slope), AGF maintains a \textbf{positive marginal gain}.}
  \label{fig:slope_analysis}
\end{figure}

A surprising finding in our experiments (Table \ref{tab:cifar100_pruning_detailed}) is that Random Pruning at $k=32$ achieves $67.79\%$, far surpassing the Training-from-Scratch baseline ($45.42\%$). 
This phenomenon suggests that the \textit{initialization values} inherited from the teacher model carry significant inductive bias. Even with a random topological subset, the network starts in a "basin of attraction" that is inaccessible from random initialization. However, our Routing analysis (Fig. \ref{fig:slope_analysis}) proves that these random sub-networks are \textbf{poorly calibrated}. While they classify well on average, their confidence scores are noisy, forcing the router to call the expensive expert more often (Cost 8.26 vs 7.28). AGF adds the necessary topological structure to maximize dynamic efficiency.

\subsection{Limitations and Boundary Conditions}
We acknowledge three limitations in this study:
\begin{itemize}
    \item \textbf{Architectural Scope:} Our initial rigorous analysis and theoretical validation naturally focus on ResNet-style CNNs, which possess strong inductive biases. However, to truly test the universality of our framework, we subsequently extend our structural probing to Vision Transformers (ViTs)—architectures notoriously lacking such spatial priors—in Part VI.
    \item \textbf{Calibration Overhead:} Unlike zero-cost static metrics ($\ell_1$), AGF requires a calibration phase (approx. 10-20 backward passes). However, this is a one-time offline cost that is negligible compared to the long-term inference savings.
    \item \textbf{Baseline Selection:} Our study is designed as an analysis of metric properties (Gradient vs. Magnitude). We therefore benchmark against canonical pruning metrics (Taylor, $\ell_1$, Random) to isolate the metric's contribution, rather than combining orthogonal engineering tricks (e.g., distillation) to chase state-of-the-art leaderboards.
\end{itemize}

\section{Conclusion}
We have presented a unified framework connecting feature learning theory (AGF) to system efficiency. By identifying the "Performance Degradation Threshold" in pruning and resolving "Gradient-Magnitude Decoupling" in routing, we demonstrate that gradient-flow utilities provide the structural foundation for next-generation dynamic networks.

\appendices

\section{Detailed Experimental Results}
\label{app:detailed_results}

\subsection{Confidence-Based Routing Numerical Sweep}
Table \ref{tab:routing_sweep_full} provides numerical results of the threshold sweep for AGF, $\ell_1$-Norm, and Random pruning strategies, as referenced in Section \ref{sec:routing_perf}.

\begin{table*}[t]
\centering
\caption{\textbf{Full Numerical Sweep of Confidence-Based Routing.} We compare the Accuracy (\%) and Normalized Cost of three strategies across varying confidence thresholds $\tau$. \textbf{Bold} indicates the best trade-off (Highest Accuracy or Lowest Cost) at critical operating points.}
\label{tab:routing_sweep_full}
\resizebox{0.9\textwidth}{!}{
\begin{tabular}{c|cc|cc|cc|c}
\toprule
\multirow{2}{*}{\textbf{Threshold ($\tau$)}} & \multicolumn{2}{c|}{\textbf{AGF (Ours)}} & \multicolumn{2}{c|}{\textbf{$\ell_1$-Norm}} & \multicolumn{2}{c|}{\textbf{Random}} & \multirow{2}{*}{\textbf{Regime}} \\
 & \textbf{Acc (\%)} & \textbf{Cost} & \textbf{Acc (\%)} & \textbf{Cost} & \textbf{Acc (\%)} & \textbf{Cost} & \\
\midrule
0.000 & 68.54 & \textbf{1.00} & \textbf{68.83} & 1.00 & 68.13 & 1.00 & \textit{Pruned Only} \\
\midrule
0.500 & 69.26 & \textbf{7.28} & \textbf{69.61} & 7.93 & 69.25 & 8.26 & \multirow{3}{*}{\textit{Low Cost}} \\
0.700 & 70.94 & \textbf{21.33} & \textbf{71.04} & 21.35 & 70.87 & 21.69 & \\
0.800 & 71.40 & \textbf{28.39} & \textbf{71.47} & 29.16 & 71.37 & 29.30 & \\
\midrule
0.900 & 72.10 & \textbf{38.08} & 72.11 & 39.29 & 72.08 & 39.61 & \multirow{3}{*}{\textit{Balanced}} \\
0.950 & 72.35 & \textbf{45.70} & \textbf{72.51} & 47.39 & 72.35 & 47.63 & \\
0.980 & 72.48 & \textbf{54.61} & \textbf{72.66} & 56.39 & 72.48 & 56.85 & \\
\midrule
0.990 & \textbf{72.65} & \textbf{60.44} & 72.57 & 62.57 & 72.48 & 62.66 & \textit{\textbf{Peak (Last Mile)}} \\
0.999 & \textbf{72.49} & \textbf{76.17} & 72.33 & 78.71 & 72.11 & 80.80 & \textit{Over-Conservative} \\
\midrule
1.000 & 71.17 & 150.41 & 71.17 & 150.41 & 71.17 & 150.41 & \textit{Full Expert} \\
\bottomrule
\end{tabular}
}
\end{table*}

\section{Implementation Details}
\label{app:implementation}

To facilitate reproducibility, we provide the specific hardware equipment and hyperparameter settings used in our experiments.

\subsection{Hardware and Software Environment}
All CIFAR-100 experiments were conducted on a workstation with the following specifications:
\begin{itemize}
    \item \textbf{CPU:} AMD Ryzen 7 7840H (8 cores, 16 threads).
    \item \textbf{GPU:} NVIDIA GeForce RTX 4060 (8GB VRAM).
    \item \textbf{Software Stack:} PyTorch 2.9.0 (Nightly/Custom Build) + CUDA 12.6.
\end{itemize}
ImageNet-100 scalability experiments were performed using NVIDIA L4/T4 GPUs via Google Colab Pro.

\subsection{Training Recipes}

\textbf{1) Teacher Model Pre-training:}
We train the WideResNet-18-2 teacher model from scratch to ensure a stable, saturated baseline.
\begin{table}[h]
\centering
\caption{Hyperparameters for Teacher Training (CIFAR-100).}
\label{tab:teacher_hyperparams}
\begin{tabular}{l|l}
\toprule
\textbf{Parameter} & \textbf{Value} \\
\midrule
Optimizer & SGD (Nesterov) \\
Momentum & 0.9 \\
Weight Decay & $5 \times 10^{-4}$ \\
Total Epochs & 150 \\
Batch Size & 32 \\
Initial Learning Rate & $1.0 \times 10^{-3}$ \\
LR Schedule & Cosine Annealing \\
Data Augmentation & RandomCrop, RandomHorizontalFlip \\
\bottomrule
\end{tabular}
\end{table}

\textbf{2) Pruning and Fine-tuning:}
For the AGF pruning phase, we employ an iterative schedule to allow the network dynamics to heal.
\begin{itemize}
    \item \textbf{Initialization:} Weights inherited from the converged Teacher (Epoch 150).
    \item \textbf{Fine-tuning LR:} Fixed at $1.0 \times 10^{-3}$ (CIFAR) and $1.0 \times 10^{-4}$ (ImageNet).
    \item \textbf{AGF Calculation:} We use $T=4 \sim 8$ batches for score accumulation.
\end{itemize}


\ifCLASSOPTIONcaptionsoff
  \newpage
\fi

\bibliographystyle{IEEEtran}
\bibliography{references}

\end{document}